\documentclass[10pt,letterpaper]{article}

\usepackage{amsthm,amsmath}

\usepackage[utf8]{inputenc} 

\usepackage{nameref,hyperref}

\usepackage{graphicx}
\usepackage{xcolor}
\usepackage{multirow}
\usepackage{xspace}
\usepackage{bold-extra}

\usepackage{tikz}
\usetikzlibrary{shapes}
\usetikzlibrary{shapes.multipart}
\usetikzlibrary{fit}

\usepackage{amssymb}
\usepackage{mathrsfs}

\usepackage{geometry}
\usepackage{pdflscape}

\usepackage{booktabs}
\usepackage{makecell}

\usepackage{todonotes}

\begin{document}
\vspace*{0.35in}

\begin{flushleft}
{\Large
\textbf\newline{Facilitating phenotyping from clinical texts: the medkit library}
}
\smallskip

Antoine Neuraz\textsuperscript{1,2,3}, 
Ghislain Vaillant\textsuperscript{1,2},
Camila Arias\textsuperscript{1,2},
Olivier Birot\textsuperscript{1,2},\\
Kim-Tam Huynh\textsuperscript{1,2},
Thibaut Fabacher\textsuperscript{1,2,4},
Alice Rogier\textsuperscript{1,2,5},
Nicolas Garcelon\textsuperscript{1,2,6},
Ivan Lerner\textsuperscript{1,2,5},
Bastien Rance\textsuperscript{1,2,5},
Adrien Coulet\textsuperscript{1,2,$\star$}\\

\bigskip

\text{\textsuperscript{1}} Inria Paris, Paris, France
\\
\text{\textsuperscript{2}} Centre de Recherche des Cordeliers, Inserm, Université Paris Cité, Sorbonne Université, Paris, France
\\
\text{\textsuperscript{3}} Hôpital Necker, Assistance Publique - Hôpitaux de Paris, Paris, France
\\
\text{\textsuperscript{4}} University Hospital of Strasbourg, Strasbourg, France
\\
\text{\textsuperscript{5}} Hôpital Européen Georges Pompidou, Assistance Publique - Hôpitaux de Paris, Paris, France 
\\
\text{\textsuperscript{6}} Imagine Institute, Inserm UMR 1163, Université Paris Cité, Paris, France 
\\

\bigskip
\textsuperscript{$\star$} corresponding author: \texttt{adrien.coulet@inria.fr}

\end{flushleft}

\bigskip
\bigskip

\abstract{Phenotyping consists in applying algorithms to identify individuals associated with a specific, potentially complex, trait or condition, typically out of a collection of Electronic Health Records (EHRs). Because a lot of the clinical information of EHRs are lying in texts, phenotyping from text takes an important role in studies that rely on the secondary use of EHRs. 
However, the heterogeneity and highly specialized aspect of both the content and form of clinical texts makes this task particularly tedious, and is the source of time and cost constraints in observational studies.  
To facilitate the development, evaluation and reproductibility of phenotyping pipelines, we developed an open-source Python library named \texttt{medkit}. It enables composing data processing pipelines made of easy-to-reuse software bricks, named medkit operations. In addition to the core of the library, we share the operations and pipelines we already developed and invite the phenotyping community for their reuse and enrichment. \texttt{medkit} is available at \url{https://github.com/medkit-lib/medkit}.\\
\textbf{keywords: }phenotyping, clinical texts, feature extraction, reproducible computing, open science}


\bigskip
\bigskip

\section{Introduction}
The collection at large scale of Electronic Health Records (EHRs) and the constitution of Clinical Data Warehouses (CDW) enable the design of clinical studies relying on a secondary use of healthcare data \cite{madigan14}. 
A substantial part of the necessary information to conduct these studies is available in texts, such as clinical notes, hospitalization or exam reports \cite{kharrazi18}. 
For instance, tasks such as the inclusion / exclusion of patients, and the extraction of outcome variables or covariates often require the consideration of clinical texts. 

In biomedical data sciences, the two complementary tasks of either identifying individuals associated with a specific, potentially complex, trait or condition, or listing the traits of an individual are generally named \textit{phenotyping}. And the specific case of phenotyping from clinical text is a continuous challenge for several reasons \cite{banda18}. First clinical text is highly specialized as it includes various factors of complexity such as medical entities absent from the general domain, hypotheses, negations, abbreviations, personal information; what motivates the development of dedicated phenotyping tools \cite{kreimeyer17}.  Besides, many powerful Natural Language Processing (NLP) tools and models are developed and shared for both the general and biomedical texts, making reuse, adaptation and chaining rational approaches in biomedicine. But the highly heterogeneous aspect of clinical texts (\textit{e.g.}, physician \textit{vs.} nurse notes, hospital A \textit{vs.} hospital B notes, French \textit{vs.} English notes) makes the performance of a tool hardly predictable on a new corpus. In addition, clinical texts can hardly be shared because of their personal and sensitive aspects. This implies the need for tools that ease the adaptation and evaluation of phenotyping approaches to the various types of texts generated in the large variety of existing clinical settings.  

We present here medkit, an open-source Python library, that aims primarily at facilitating the reuse, evaluation, adaptation and chaining of NLP tools for the development of reproducible phenotyping pipelines. By extension, medkit enables the extraction of information related to patient care, such as treatments or procedures, which are not phenotype per se. The rest of this manuscript presents the core elements of the library, details example pipelines developed with medkit for the extraction of drug treatments from clinical texts, and lists other operations and pipelines already developed and ready for reuse. It ends on two particularity added values of the library, which are the support of non-destructive processing and provenance tracing.

\section{Related work and positioning}\label{sec2}

The PheKB initiative proposes a collaborative web portal to share phenotyping algorithms in the form of semi-formal flow charts, documenting their steps and chaining \cite{kirby2016}. PheKB helps exchanging and standardizing phenotyping algorithms, however those are independent from their computational implementations, therefore limiting their reproducibility and comparison. In addition, algorithmic steps that rely on clinical texts are underspecified, as they usually require an adaptation to the peculiarities of local texts. 
The OHDSI community offers software tools such as Atlas, which proposes standard and reusable tools for the data analysis of observational studies from EHRs \cite{ohdsi2019book}. Those are developed for steps coming next to the information extraction, once features are structured and normalized. medkit fills this exact step, extracting and normalizing features from unstructured parts of EHRs. 
The MedCAT library focuses on entity recognition and normalization with the UMLS \cite{Kraljevic2021-ln}. The 
Gate suite provides a graphical user interface which facilitates sequential application of various preprocessing and NLP tools on texts \cite{cunningham2002gate}. Gate is developed in Java and is mostly adapted to educational or exploratory purposes, but has limited capabilities in analysis of large corpora and in ease of reuse of external tools. 
NLTK (Natural Language Toolkit) \cite{bird2009natural}, spaCy \cite{spacy2} and FLAIR \cite{akbik2019flair} are Python libraries dedicated to advanced NLP development, designed for NLP engineers and researchers. medkit aims at being easier to start with, facilitating the reuse and chaining of simple-to-complex NLP tools, such as those developped with the previously cited libraries. 

One of the main particularity of medkit is to place a strong emphasis on non-destructive operations, \textit{i.e.}, no information is lost when passing data from one step to another; and on a flexible tracing of data provenance.
In this matter, medkit is original and found inspiration in bioinformatics workflow management systems, such as Galaxy and Snakemake \cite{galaxy2022, molder2021}, which facilitate 
reproducibility of bioinformatics pipelines.

\section{The core components of medkit}\label{sec3}

For internal data management, medkit represents data with three simple core classes: Documents, Annotations and Attributes. 
Each of these classes is associated with properties and methods to represent data and metadata of various modalities such as audio or images, even though medkit is primarily designed for text. Document defines the minimal data structure of medkit, which associates an identifier and a set of Annotations; in turn each Annotation associates an identifier, a label and a set of Attributes; lastly each Attribute associates an identifier, a label and a value. 

For data processing, medkit defines two main classes: Operations and Pipelines. Typically, an operation is taking data as an input, runs a function over these data and returns output data. For instance an Operation can input a Document, perform Named Entity Recognition (NER) and output a set of Annotations associated with the Document. Accordingly, an operation can be the encapsulation of a previously developed tool,  or a new piece of software developed in Python using medkit classes. Converters are particular operations for input and output management, which enable the transformation from standard formats such as CSV, JSON, Brat, Docanno annotations, into medkit Documents, Annotations and Attributes, or inversely. 
Lastly, Pipelines enable to chain Operations within processing workflows. 

We refer to the medkit documentation for more details on its core components (see Availability Section for a web link). 

\section{Encapsulate, chain, and reuse}
Numerous data processing tools exist, in particular in NLP, where pretrained models are routinely shared within libraries or platforms such as spaCy or Hugging Face~\cite{spacy2,wolf2020huggingfaces}. The goal of medkit is to facilitate their reuse, evaluation and chaining. Following are examples of  available medkit operations that reuse third-party tools: the Microsoft library named Presidio for text de-identification~\cite{MsPresidio}; a date and time matcher from the EDS-NLP lib\cite{edsnlp}; text translator using transformers from the Hugging Face platform. Similarly, medkit operations enable the encapsulation of spaCy modules in particular by input, output and annotation conversion functions. We aim at progressively enriching the catalogue of tools, thanks to the continuous growth of the community of medkit users and contributors.

\section{Example pipelines}\label{sec4}

\begin{figure}[!t]%
\centering
\includegraphics[width=7.8cm]{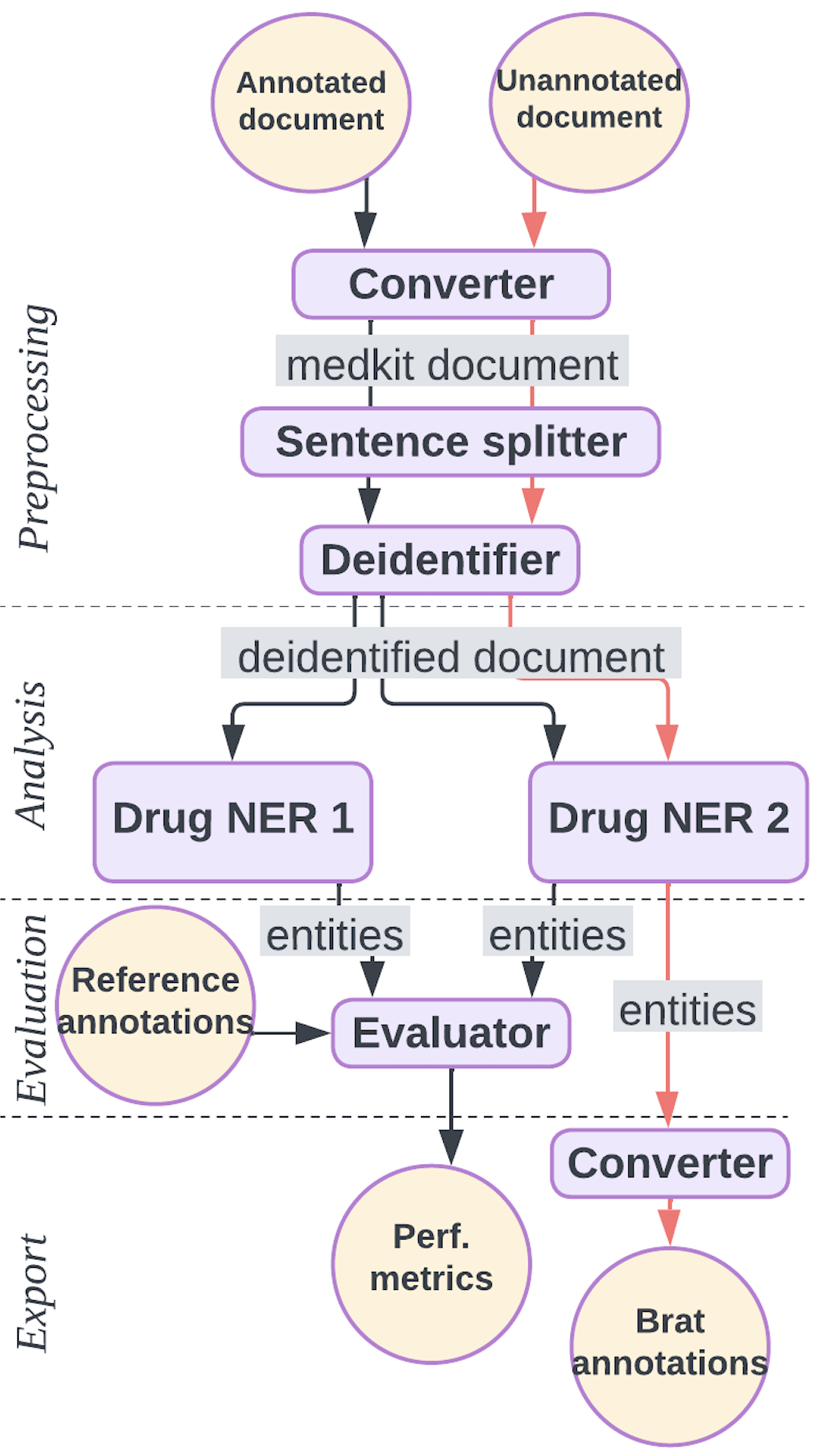}
\caption{Example medkit pipelines. The black pipeline converts raw texts to the medkit format, deidentifies them, recognizes drug entities with two distinct tools and compute performances for comparison. The orange pipeline, performs the same preprocessing tasks, recognizes drugs with only Drug NER 2 and outputs annotations in the Brat format.}\label{fig:fig1}
\end{figure}

\begin{figure}[!t]%
\centering
\includegraphics[width=8.4cm]{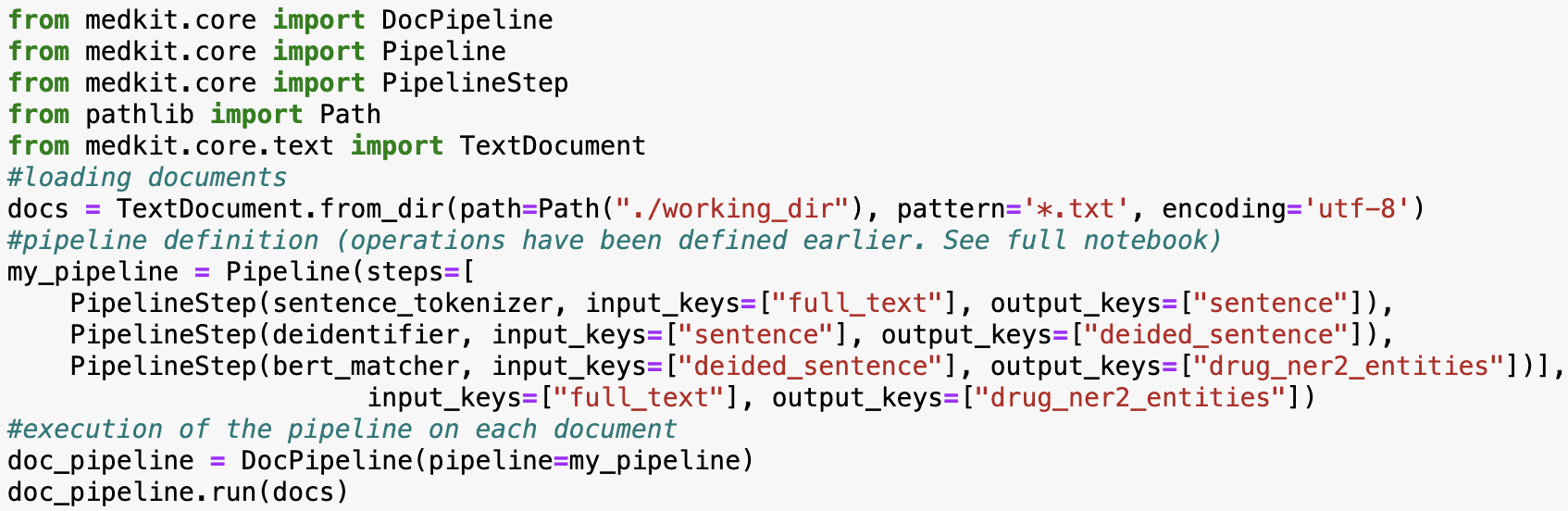}
\caption{Snippet of the implementation of the orange pipeline of Fig. \ref{fig:fig1}. }\label{fig:fig2}
\end{figure}

As an illustration, we describe two medkit pipelines in Figure \ref{fig:fig1} for the extraction of drug treatment from clinical text. The first pipeline, in black, aims at comparing the performances of two NER tools named Drug NER 1 and 2, which are dictionary-based and Transformer-based methods respectively. Considering that Drug NER 2 obtained the best performances, the second pipeline is designed to use only the latter to extract the mentions of drug treatments from new texts. Both pipelines share three steps of preprocessing: convertion of raw texts into medkit documents, sentence splitting and de-identification. 
The first pipeline evaluates the two tools on the basis of reference annotations saved as Brat format, whereas the second pipeline annotates new documents with drugs and produces output annotations in Brat format. A snippet of code for the medkit implementation of the second pipeline is shown in Fig. \ref{fig:fig2}. The full implementation of the two pipelines is available at  
{\small \url{https://medkit.readthedocs.io/en/latest/cookbook/drug_ner_eval/}}.

\section{Available operations and pipelines}\label{sec5}

We developed and share operations for: NER; relation extraction; preprocessing; deidentification; evaluation; the pre-annotation of clinical texts to speed-up manual annotation; the detection of negation, hypothesis and antecedents within the context of entities; the fine-tuning of preexisting models; the loading of audio patient-caregiver conversations, their diarisation and transcription to text.

We implemented and share pipelines for: the phenotyping of chemotherapy toxicities, and their grades \cite{rogier:hal-03364585}; the phenotyping of rheumatoid arthritis in French clinical reports \cite{fabacher:hal-04069779}; the phenotyping of COVID-19 and the comparisons of pipelines relying either on the English \textit{vs.} French UMLS \cite{neuraz:hal-04069590}; the benchmarking of NER approaches on three clinical case corpora, comparing dictionary-based, transformer and generative approaches \cite{hubert24}; the detection of text duplications in collections of clinical texts \cite{fabacher_2023}.

We refer the reader to the tutorial and cookbook sections of the medkit documentation for a list of available operations and  examples of pipelines (see Availability Section for a web link).

\section{Non-destructive processing and provenance}\label{sec6}

The medkit library ensures two uncommon functionalities: non-destructive processing and flexible provenance tracing. 
Non-destructive processing ensures that no information is lost when passing from one operation to the next. This is of particular interest when one wants to visualize annotations on a raw text, after this one underwent various transformating steps, such as deidentification or character replacements. Those change the text, the relative position of words in term of character offset, making such visualization challenging. 
Non-destructive processing is enabled by the propagation of original spans through successive operations. We note that this functionality might be lost in the case of external and noncompliant tools encapsulated in medkit operations. 

Provenance tracing consists in recording provenance data, \textit{i.e.}, meta-data documenting where data come from and how it was transformed~\cite{bose2005}. medkit implements this tracing by generating provenance data using the PROV-O standard ontology \cite{lebo2013prov}. This tracing is flexible in the sense that users can set the level of verbosity and details they want to trace about the previous operations and states, in order to avoid generating large amounts of provenance data when those are unnecessary. 

The unique combination of non-destructive processing and provenance tracing improves the explainability and reproducibility of results of pipelines of various level of complexity. These functionalities, along with the open source nature of medkit and its focus on interoperability with existing libraries, pipelines and models, make it well aligned with the FAIR principles for research software \cite{barker_introducing_2022}.

\section{Availability}\label{sec7}

medkit is at \url{https://github.com/medkit-lib/medkit}, and released under an MIT license. Its documentation, with examples and tutorials, is at \url{https://medkit.readthedocs.io/}.

\section{Conclusion and perspectives}

medkit is an open source library for the composition of data processing pipelines made of easy-to-reuse software bricks, which aim at facilitating phenotyping from clinical texts. In addition to the core of the library, we share many of these bricks and examples of pipelines, and invite the phenotyping community for their reuse and enrichment.

So far, medkit enables linear execution of pipelines over a set of documents. Whereas it is simple to distribute the execution of pipelines by splitting a large corpus in subsets, parallelization within pipelines or operations is not supported yet. It is however one feature we would like to add soon to medkit. 
We would like to grow the community of users of medkit, first by developing a searchable library of the available operations, by enriching this library and enabling users to share their own pipelines. Pipelines may be showcased in a gallery of examples to inspire and facilitate reuse.
This effort would require the formalization of a process for contributors to submit their proposals for new operations, and for maintainers to validate the quality of those submissions. 
The next operations we will develop concern the normalization capabilities of the library, the generation of features that are compliant with the OMOP Common Data Model, and operations that  facilitate the use of large language models and prompting. 















\section{Competing interests}
No competing interest to declare.

\section{Author contributions statement}

G.V., C.A., O.B. and K.T.H. designed and implemented the library and reviewed the manuscript. T.F. and A.R. implemented pipelines, discussed use cases and reviewed the manuscript. A.N., N.G., I.L., B.R. and A.C. obtained the funding, supervised the development, participated in the design, wrote and reviewed the manuscript.

\section{Acknowledgments}
Authors thank users of medkit, in particular L.-A. Guiottel, M. Hassani, S. Cossin, T. Hubert, V. Pohyer for their insightful inputs. This work was supported by the Digital Health Program of Inria; Inria Paris; and the Agence Nationale de la Recherche under the France 2030 program [ANR-22-PESN-0007].

\bibliographystyle{unsrt}
\bibliography{reference}

\end{document}